\documentclass[10pt,twocolumn,letterpaper]{article}

\usepackage{cvpr}              
\usepackage{multirow}

\definecolor{cvprblue}{rgb}{0.21,0.49,0.74}
\usepackage[pagebackref,breaklinks,colorlinks,allcolors=cvprblue]{hyperref}

\def\paperID{2721} 
\def\confName{CVPR}
\def\confYear{2026}

\title{AEGPO: Adaptive Entropy-Guided Policy Optimization for Diffusion Models}

\author{
    Yuming Li$^{1,2}$\textsuperscript{*}, 
    Qingyu Li$^{2}$\textsuperscript{*}, 
    Chengyu Bai$^{1}$\textsuperscript{*}, 
    Xiangyang Luo$^{2}$, 
    Zeyue Xue$^{3}$, 
    Wenyu Qin$^{2}$, \\
    Meng Wang$^{2}$, 
    Yikai Wang$^{4}$\textsuperscript{\ddag}, 
    Shanghang Zhang$^{1}$\textsuperscript{\ddag} \\
    $^1$Peking University, Beijing, China \quad
    $^2$Kling Team, Kuaishou Technology, Beijing, China \\
    $^3$The University of Hong Kong, Hong Kong \quad
    $^4$Beijing Normal University, Beijing, China \\
    \small
    \textsuperscript{*}Equal contribution. \quad 
    \textsuperscript{\dag}This work was conducted during the author's internship at Kling Team, Kuaishou Technology. \quad 
    \textsuperscript{\ddag}Corresponding author.
}

\begin{document}
\maketitle
\begin{abstract}
Reinforcement learning from human feedback (RLHF) shows promise for aligning diffusion and flow models, yet policy optimization methods like GRPO suffer from inefficient, static sampling strategies. These methods treat all prompts and denoising steps uniformly, overlooking significant variations in sample learning value and the {dynamic nature of critical exploration moments} .
To address this, we first conducted a detailed analysis of the internal attention dynamics during GRPO training, unveiling a critical insight: Attention Entropy serves as a powerful, dual-signal proxy. (1) Across different samples, we found that the relative Attention Entropy change ($\Delta$Entropy), the divergence between the current and base policy, acts as a robust proxy for {sample learning value}. (2) During the denoising process, we find that the {peaks} of the absolute Attention Entropy ($Entropy(t)$)—which quantifies {attention dispersion}—effectively identify {critical, high-value exploration timesteps}.
Based on this core discovery, we propose {Adaptive Entropy-Guided Policy Optimization (AEGPO)}, a novel dual-signal, dual-level adaptive strategy. \textbf{AEGPO} leverages these intrinsic signals: (1) Globally, it uses $\Delta$Entropy to dynamically allocate rollout budgets, prioritizing high-value prompts. (2) Locally, it uses the {peaks} of $Entropy(t)$ to guide exploration only at these {critical high-dispersion timesteps} .
By concentrating computation on the most informative samples and moments, \textbf{AEGPO} enables more effective policy optimization. Experiments on text-to-image generation demonstrate that \textbf{AEGPO} significantly accelerates convergence and achieves superior alignment compared to standard GRPO variants.
\end{abstract}

\section{Introduction}
Diffusion models and flow-matching models have achieved state-of-the-art performance in visual synthesis~\cite{ho2020ddpm,lipman2022flow,liu2022flow}. To align these powerful models with human preferences, Reinforcement Learning from Human Feedback (RLHF), and in particular Group Relative Policy Optimization (GRPO), has emerged as a stable and scalable optimization paradigm~\cite{ouyang2022training}. Recent extensions have applied GRPO to diverse domains—text-to-image generation, video synthesis, aesthetic alignment, and preference optimization~\cite{liu2025flow,xue2025dancegrpo,he2025tempflow,li2025branchgrpo,zheng2025diffusionnft,li2025mixgrpo,fu2025dynamic,wang2025grpo,yu2025smart,zhou2025text,lee2025pcpo}.

Despite the rapid adoption of GRPO variants, existing approaches exhibit two fundamental inefficiencies. \textbf{First, static rollout allocation} treats all prompts equally, assigning a uniform number of rollouts per sample. This ignores the substantial variation in how much different prompts contribute to policy improvement. \textbf{Second, timestep-level exploration is often governed by fixed, predefined schedules} or expensive reward-based attribution. Such methods cannot adapt to the highly dynamic nature of the denoising process and policy optimization process, where critical denoising steps are varying for different prompts.

\begin{figure}
    \centering
    \includegraphics[width=0.9\linewidth]{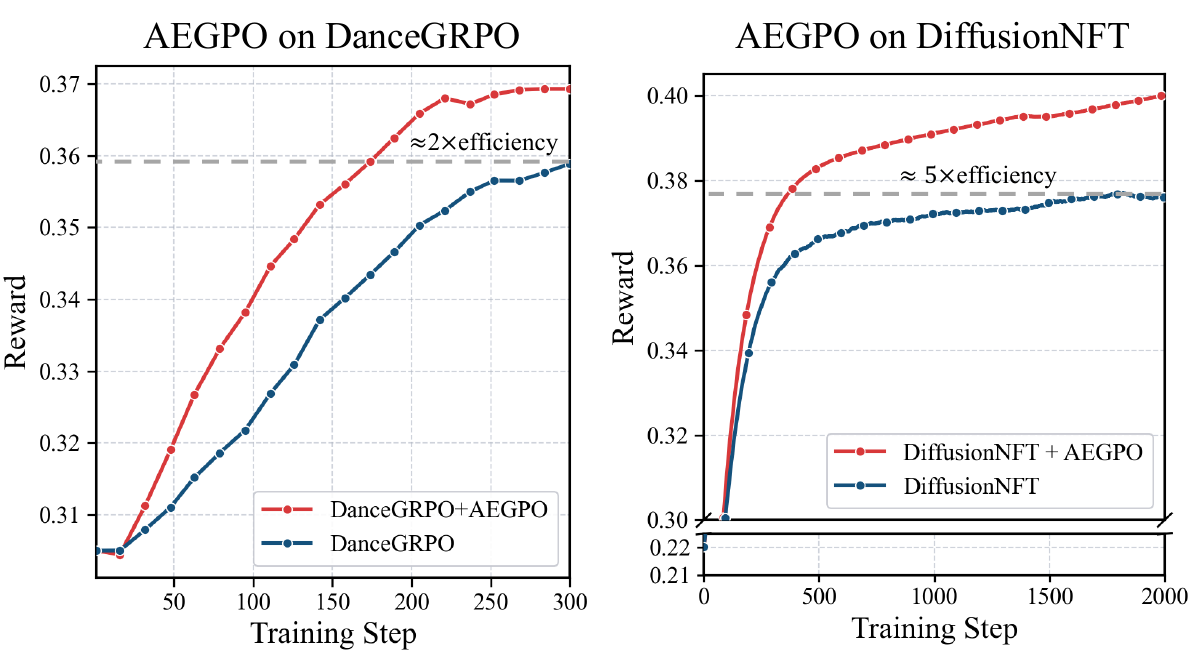}
    \caption{AEGPO significantly accelerates policy optimization. Compared to standard GRPO variants, AEGPO achieves 2$\times$ faster convergence on DanceGRPO (left) and 5$\times$ faster convergence on DiffusionNFT (right), while also reaching a superior final reward.}    
    \label{fig:teaser}
\end{figure}

These limitations raise two essential questions for efficient policy optimization:  
(1) \emph{How can rollout budgets be allocated dynamically according to each prompt's learning value?}  
(2) \emph{How can we identify critical timesteps for exploration using an intrinsic signal, without relying on costly reward attribution?}

\begin{figure*}[t]
    \centering
    \includegraphics[width=\linewidth]{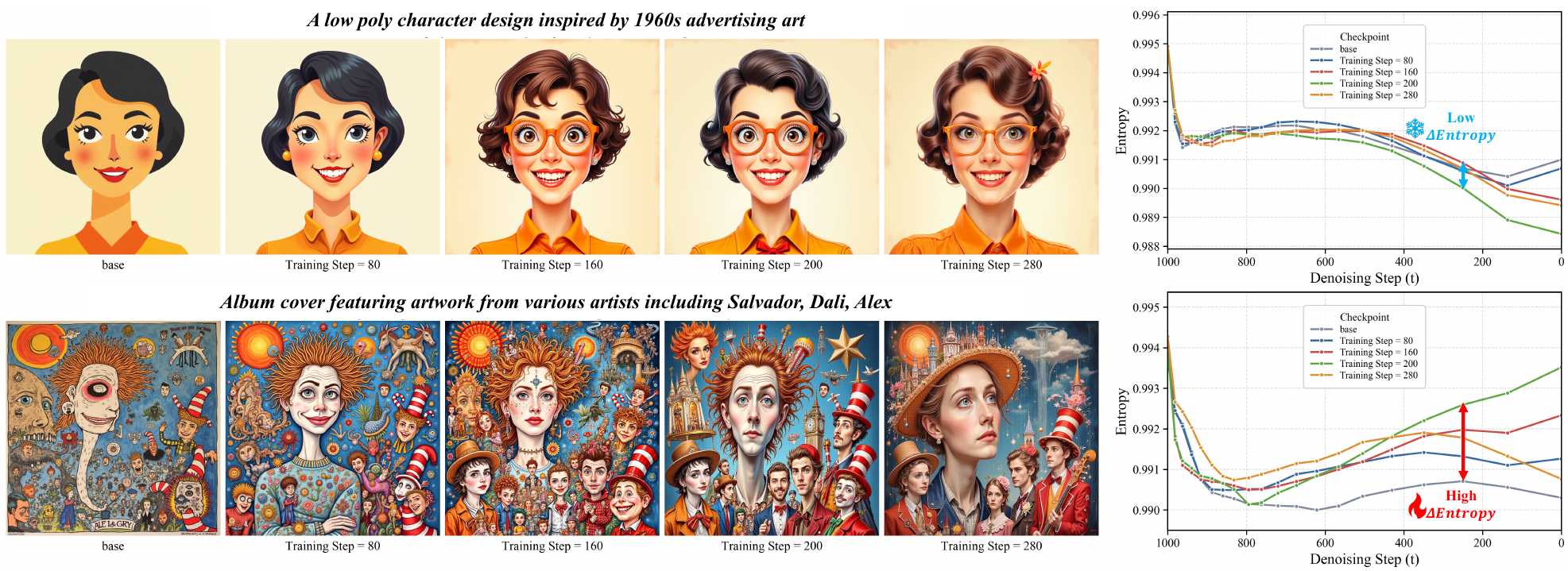}
    \caption{
        Illustration of varied Attention Entropy dynamics during GRPO training. (Left Panels): Generated images evolving across training steps. The top prompt shows minor visual changes, while the bottom prompt undergoes significant improvement. (Right Panels): The corresponding Absolute Attention Entropy ($Entropy(t)$) trajectories over the denoising steps $t$. The entropy curves for the top prompt remain clustered, indicating a low overall Relative Entropy Change ($\Delta$Entropy). Conversely, those for the bottom prompt show substantial divergence (indicating a high overall $\Delta$Entropy), which correlates with the degree of visual change. \textit{We provide additional qualitative visualizations, as well as a detailed analysis of the impact of different reward models on entropy dynamics, in the Appendix.}
    }
    \label{fig:intro_teaser} 
\end{figure*}

To answer these questions, we perform a detailed study of attention entropy during GRPO training. Our analysis reveals two intrinsic signals that naturally address the challenges above. \textbf{At the sample level,} changes in attention behavior between the current policy and the base policy provide a strong indicator of a sample's learning value. \textbf{At the timestep level,} the magnitude of attention dispersion highlights the most informative decision points within each rollout. These insights indicate that both sample value and critical exploration timestep vary dramatically across different prompts and different policy optimization stage, making static heuristics incapable of adapting to them.

Building on the analysis of attention entropy, we propose \textbf{Adaptive Entropy-Guided Policy Optimization (AEGPO)}, a plug-and-play framework for improving rollout efficiency. AEGPO introduces two complementary adaptive strategies:  
(1) \emph{Global Adaptive Allocation}, which assigns rollout budgets based on sample learning value, concentrating computation on high-impact prompts;  
(2) \emph{Local Adaptive Exploration}, which identifies critical timesteps by detecting peaks in the model’s attention dispersion.  

We validate AEGPO across diverse policy optimization frameworks, including DanceGRPO~\cite{xue2025dancegrpo}, BranchGRPO~\cite{li2025branchgrpo}, FlowGRPO~\cite{liu2025flow}, and DiffusionNFT~\cite{zheng2025diffusionnft}, and across multiple base models such as FLUX.1-dev~\cite{flux2024} and SD3.5-M~\cite{esser2024scaling}. AEGPO consistently accelerates convergence (up to $5\times$ faster) and improves final reward.

{Our contributions are as follows:}
\begin{itemize}
\item We establish Attention Entropy as a dual-signal intrinsic proxy, where relative entropy change ($\Delta$Entropy) indicates sample learning value and absolute entropy peaks ($Entropy(t)$) identify critical exploration timesteps.
\item We introduce AEGPO, a dual-level adaptive framework that improves sample selection and timestep exploration using lightweight entropy-guided mechanisms.
\item We demonstrate that AEGPO is general, efficient, and broadly applicable, achieving significant improvements across multiple policy optimization methods and diffusion backbones.
\end{itemize}

\section{Related Work}
\label{sec:related_work}

Aligning diffusion and flow models with human preferences is an active area of research . Early approaches adapted existing RL paradigms. Diffusion-DPO \cite{wallace2024diffusion, rafailov2023direct} extends DPO to diffusion models, directly optimizing on preference pairs. Policy gradient methods like DDPO \cite{black2023training} apply REINFORCE-like algorithms but often face high variance. Another line leverages differentiable reward models to backpropagate reward signals, but these methods can be computationally intensive and sensitive to the choice of optimization steps . Our work builds on the RLHF framework, but instead of proposing a new loss, we focus on improving the sampling efficiency of policy optimization algorithms like GRPO.

Group Relative Policy Optimization (GRPO) \cite{xue2025dancegrpo, liu2025flow} offers a stable and effective RLHF framework well-suited for large generative models. Several works have adapted it for diffusion models: Flow-GRPO \cite{liu2025flow} and DanceGRPO \cite{xue2025dancegrpo} pioneered its application to visual generation, but typically employ uniform sampling. TempFlow-GRPO \cite{he2025tempflow} introduced noise-aware weighting \cite{he2025tempflow}, and BranchGRPO \cite{li2025branchgrpo} used structured branching with shared prefixes. SRPO \cite{shen2025directly} proposed aligning the full trajectory using interpolation. While these variants improve upon vanilla GRPO, {they often rely on external heuristics (like noise schedules) or uniform exploration strategies} that do not fully adapt to the model's internal state. AEGPO differentiates itself by using an \textit{intrinsic signal}—attention entropy—to adaptively guide \textit{both} sample allocation and exploration.

Leveraging intrinsic properties of transformer models, such as entropy, to guide learning has shown promise, particularly in LLMs \cite{shao2024deepseekmath}. For instance, works have used entropy to quantify uncertainty (SEED-GRPO \cite{chen2025seed}), balance rollouts (ARPO \cite{lu2025arpo}, AEPO \cite{dong2025agentic}), or identify critical steps for branching (AttnRL \cite{liu2025attention}) . These studies highlight the potential of using internal signals for efficiency. \textbf{AEGPO} draws inspiration from this line of research , adapting the concept of using intrinsic signals to the domain of diffusion model alignment. 

\begin{figure}[t]
    \centering
    \includegraphics[width=1\linewidth]{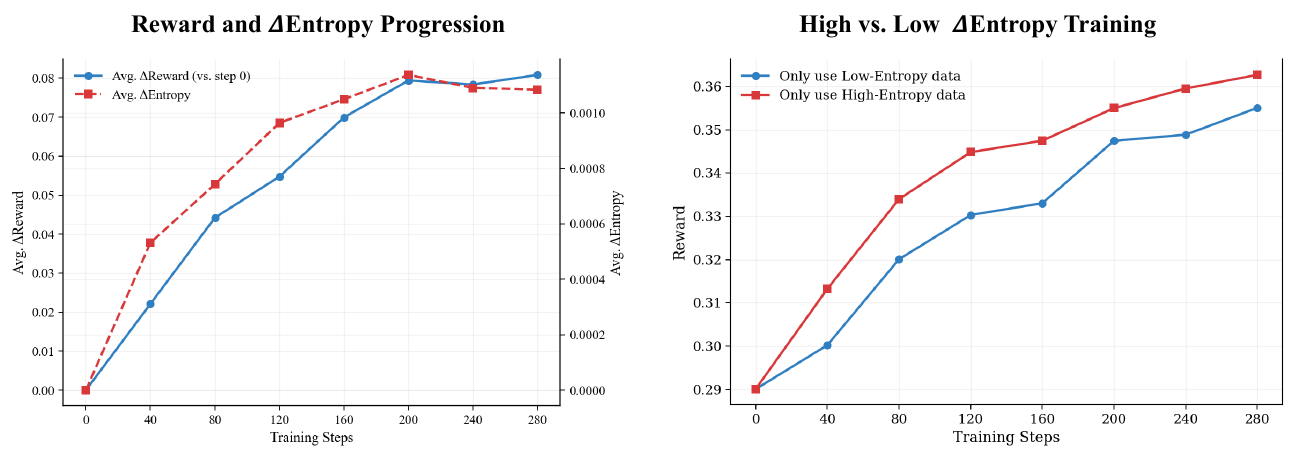}
    \caption{Validation of relative Attention Entropy change ($\Delta$Entropy) as a robust proxy for sample learning value. 
    \textbf{(Left):} In early training, both $\Delta$Reward and $\Delta$Entropy rise, indicating active policy improvement accompanied by large adjustments in attention behavior. In later stages, $\Delta$Reward plateaus and $\Delta$Entropy correspondingly stabilizes or slightly decreases, reflecting that the model has reached a confident and stable attention configuration and no longer requires large policy deviations to obtain reward gains.
    \textbf{(Right):} Reward convergence comparison. The red line shows a model trained \textit{only} on high-$\Delta$Entropy data, while the blue line shows a model trained \textit{only} on low-$\Delta$Entropy data. Training on high-value samples leads to significantly faster convergence and a superior final reward, confirming their greater learning value.}
    \label{fig:global_analysis_plots}

\end{figure}

\section{Unveiling the Attention Entropy}
In this section, we define the core intrinsic signals used in \textbf{AEGPO}: relative Attention Entropy change ($\Delta$Entropy) and absolute Attention Entropy ($Entropy(t)$). We detail their calculation and analyze their dynamics during GRPO training, providing the empirical foundation for our dual-level adaptive method.

\subsection{Defining Attention Entropy}
\label{sec:attention_definition}

We begin by formalizing the attention entropy signals used throughout our empirical analysis. 
At each timestep $t$, the model produces an attention map $A_t$ that captures how $N$ image features $q_i$ attend to the $T$ text tokens $\tau$. 
This map is computed as:

\begin{equation}
A_t = \mathrm{softmax}\left( \frac{QK^{\top}}{\sqrt{d_k}} \right),
\label{eq:attn_map}
\end{equation}

where $Q$ and $K$ are the query and key representations. 
We average $A_t$ across the selected layers and heads to obtain $\bar{A}_t$. 
For each image feature $q_i$, its attention weights across text tokens are normalized into a probability distribution:

\begin{equation}
\mathrm{Prob}_t[q_i, \tau]
= 
\frac{\bar{A}_t[i, \tau]}
     {\sum_{\tau'=1}^{T} \bar{A}_t[i, \tau']}.
\label{eq:prob_dist}
\end{equation}

The {per-feature attention entropy} is then defined using the Shannon entropy:

\begin{equation}
\mathrm{Entropy}_t[q_i]
=
-\sum_{\tau=1}^{T}
\mathrm{Prob}_t[q_i, \tau]
\cdot
\log_2\left(\mathrm{Prob}_t[q_i, \tau]\right).
\label{eq:shannon_entropy}
\end{equation}

To characterize the model’s average attention dispersion at timestep $t$, 
we take the mean across all $N$ image features:

\begin{equation}
\mathrm{Entropy}(t)
=
\frac{1}{N}
\sum_{i=1}^{N}
\mathrm{Entropy}_t[q_i].
\label{eq:entropy_t}
\end{equation}

A higher $\mathrm{Entropy}(t)$ indicates more \textit{dispersed} attention over the text tokens,
while a lower value indicates a more \textit{focused} distribution.

To compare how the attention behavior shifts during training, 
we define the {per-timestep Relative Entropy Change} between 
the current policy $\theta$ and the base policy $\theta_{\mathrm{ba se}}$:

\begin{equation}
\Delta\mathrm{Entropy}(t)
=
\left|
\mathrm{Entropy}_{\theta}(t)
-
\mathrm{Entropy}_{\mathrm{base}}(t)
\right|.
\label{eq:delta_entropy_t}
\end{equation}

Finally, we summarize the overall {sample-level deviation} by averaging over all denoising steps $T$:

\begin{equation}
\Delta\mathrm{Entropy}
=
\frac{1}{T}
\sum_{t=1}^{T}
\Delta\mathrm{Entropy}(t).
\label{eq:delta_entropy_sample}
\end{equation}

In the following sections, we analyze $\Delta\mathrm{Entropy}$ as a \textit{sample-level} signal (Sec.~\ref{sec:global_insight}) 
and $\mathrm{Entropy}(t)$ as a \textit{timestep-level} signal (Sec.~\ref{sec:local_insight}).

\begin{figure}[t]
    \centering
    \includegraphics[width=0.8\linewidth]{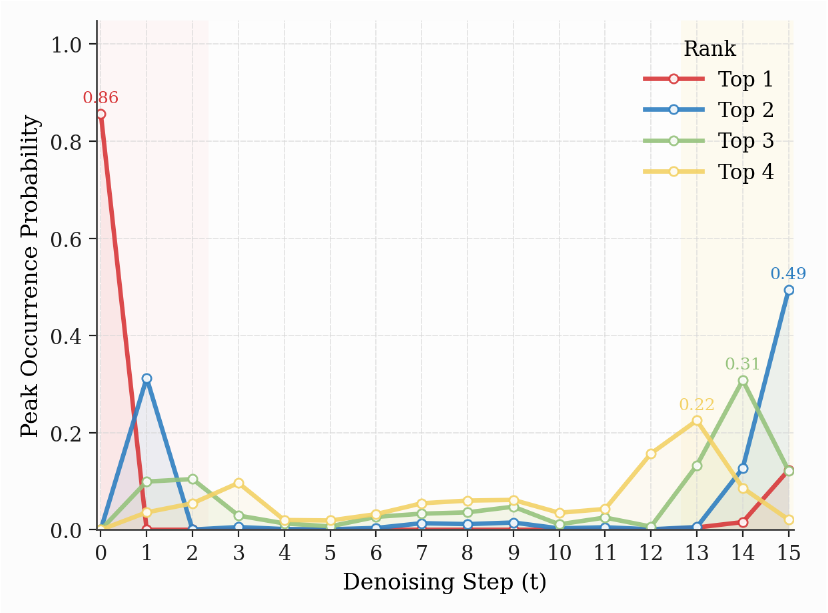}
    \caption{Distribution of Top-K Absolute Attention Entropy ($Entropy(t)$) peaks across denoising steps $t$. The y-axis shows the probability that a given step $t$ contains one of the Top-K highest entropy peaks. The distribution is distinctly U-shaped, with high-dispersion peaks clustering in the very early (e.g., $t\approx1)$ and late (e.g., $t\approx13-15)$ stages of the denoising process.}
    \label{fig:local_peak_distribution}
\end{figure}

\subsection{$\Delta$Entropy as an Indicator of Sample Value}
\label{sec:global_insight}

As defined in Sec.~\ref{sec:attention_definition}, $\Delta$Entropy measures the magnitude of change in attention dispersion between the current policy and its base counterpart. Empirically, we find that this quantity reliably reflects how much a sample \emph{forces} the model to adjust its internal attention strategy during GRPO updates, and therefore serves as an intrinsic indicator of its learning value.

This change is highly non-uniform across prompts. As illustrated in Figure~\ref{fig:intro_teaser}, prompts that induce only minor visual improvement (top row) show tightly clustered $Entropy(t)$ trajectories, indicating that the model’s attention pattern remains almost unchanged from the base policy. This suggests that the model already handles these prompts with high confidence, requiring little additional learning, and consequently yields a low $\Delta$Entropy. In contrast, prompts that trigger substantial visual improvement (bottom row) exhibit noticeably diverging $Entropy(t)$ trajectories. Here the model must meaningfully reorganize its attention allocation, signaling lower prior competence and thus a higher learning demand—captured as a large $\Delta$Entropy.

This phenomenon appears consistently at scale. In Figure~\ref{fig:global_analysis_plots} (Left), we observe a strong positive correlation between the average $\Delta$reward and the average $\Delta$Entropy across samples, indicating that larger reward gains tend to coincide with larger shifts in attention behavior. In other words, effective GRPO updates are tightly coupled with how much the policy deviates from the base model.

We further verify this by training on subsets stratified by $\Delta$Entropy (Figure~\ref{fig:global_analysis_plots}, Right). Models trained exclusively on high-$\Delta$Entropy samples converge substantially faster and reach a higher final reward than those trained on low-$\Delta$Entropy samples. This confirms that samples inducing larger attention adjustments contribute disproportionately to policy improvement.

Taken together, these findings show that $\Delta$Entropy provides a stable and informative measure of sample-level learning value, capturing how strongly a sample compels the model to revise its attention strategy during training.

\begin{table}[t]
  \centering
  \small
  \caption{Comparison of exploration value (Avg. Reward Std) and generative diversity (LPIPS MPD, TCE) achieved by fixed branching schedules versus our dynamic entropy-guided strategy, averaged over 1000 prompts from HPDv2.1. LPIPS MPD \cite{kim2025test} measures intra-prompt diversity , while TCE \cite{kim2025test} measures global diversity based on CLIP features. Higher values are better for all metrics.}
  \label{tab:reward_std_comparison}
  \begin{tabular}{@{}lccc@{}}
    \toprule
    Branching Strategy &Reward Std $\uparrow$ & LPIPS MPD $\uparrow$ & TCE $\uparrow$ \\
    \midrule
    \multicolumn{4}{l}{\textit{Fixed Schedules}} \\
    steps (0, 2, 4, 8) & 17.04 & 0.719 & 4.44 \\
    steps (0, 3, 6, 9) & 16.97 & 0.713 & 4.42 \\
    steps (0, 4, 8, 12) & 16.82 & 0.698 & 4.39 \\
    steps (0, 5, 10, 15) & 16.93 & 0.704 & 4.38 \\
    \midrule
    \multicolumn{4}{l}{\textit{Dynamic Strategy (Ours)}} \\
    \textbf{Entropy-Guided} & \textbf{17.35} & \textbf{0.727} & \textbf{4.47} \\
    \bottomrule
  \end{tabular}
\end{table}

\begin{figure*}[ht!]
    \centering
    \includegraphics[width=0.95\linewidth]{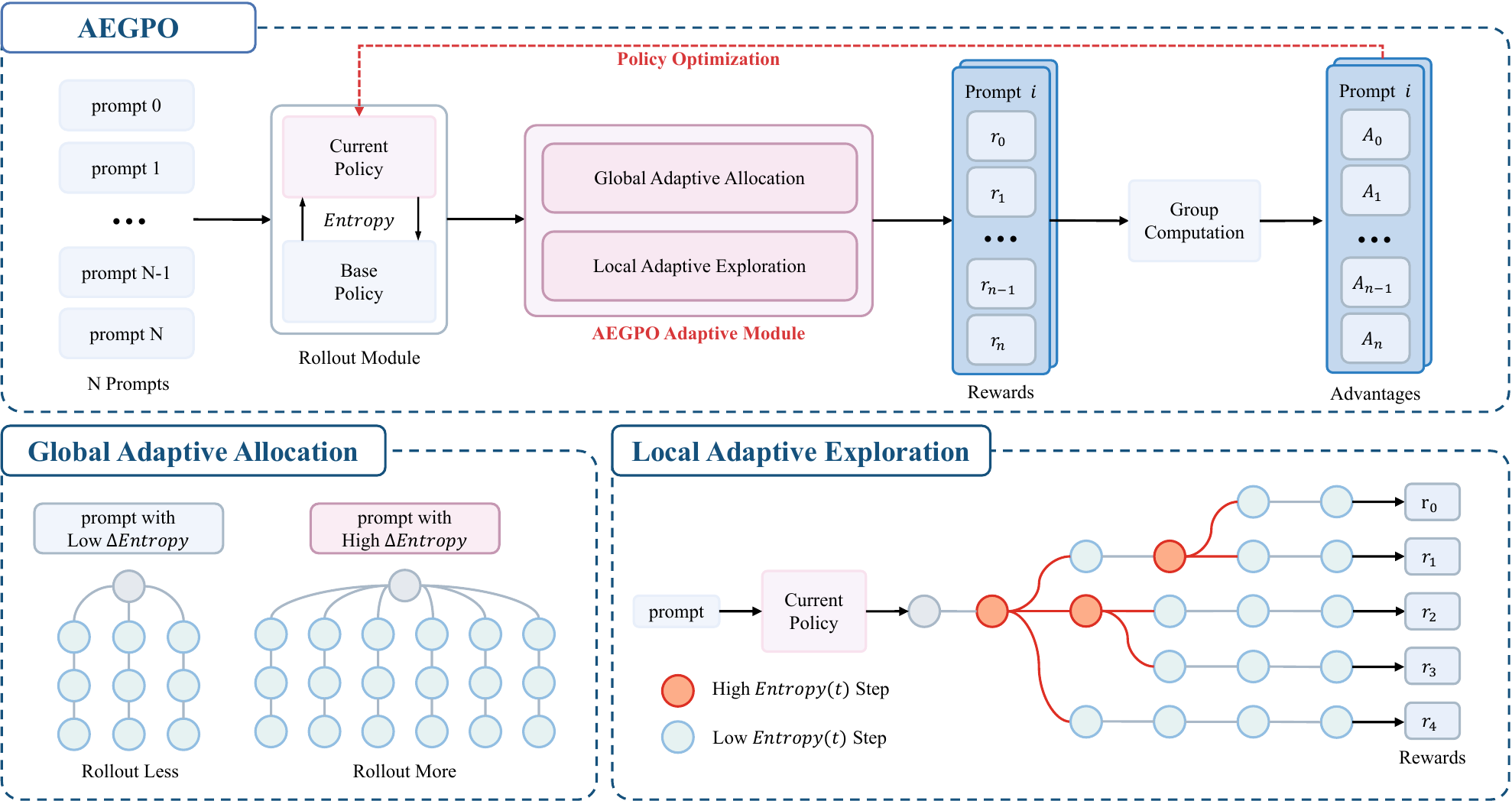}
    \caption{Overview of the \textbf{AEGPO} framework, illustrating our dual-level adaptive strategy. 
    \textbf{(Top)} The central \textbf{AEGPO Adaptive Module} (pink box) receives intrinsic signals derived from the model's policies. It uses these signals to guide the GRPO rollout and computation process.
    \textbf{(Bottom Left) Global Adaptive Allocation:} The per-sample \textbf{$\Delta$Entropy} (relative entropy change) is used as a proxy for sample value. High-value prompts are dynamically allocated a larger rollout budget (Rollout More), while low-value prompts receive fewer (Rollout Less).
    \textbf{(Bottom Right) Local Adaptive Exploration:} The per-timestep \textbf{$Entropy(t)$} (absolute entropy) is used as a proxy for model dispersion. Exploration is dynamically triggered \textit{only} at high-dispersion timesteps (orange circles), focusing exploration on the most critical moments of the denoising process.}
    \label{fig:framework_pipeline}
\end{figure*}

\subsection{$Entropy(t)$ Peaks as Indicators of Critical Timesteps}
\label{sec:local_insight}

Complementing the sample-level analysis in Sec.~\ref{sec:global_insight}, we examine the timestep-level behavior of the absolute entropy $Entropy(t)$. Large values of $Entropy(t)$ represent moments where the model distributes attention broadly across the text tokens, suggesting the need to integrate multiple semantic cues. As illustrated in Figure~\ref{fig:intro_teaser}, different prompts exhibit noticeably distinct entropy trajectories and peak patterns, indicating that the attention-dispersion behavior varies
substantially across samples.

As shown in Figure~\ref{fig:local_peak_distribution}, the distribution of peaks is not uniform; instead, they exhibit a consistent bimodal pattern. One cluster appears early, corresponding to coarse structural decisions, while another emerges late, during fine-grained refinement. Furthermore, the peak locations vary across prompts and differ across the Top~1, 2, and 3 peaks, revealing that the most uncertain timesteps are both sample-dependent and stage-dependent. This variability suggests that any fixed exploration schedule is fundamentally mismatched to the underlying dynamics.

We then evaluate whether these high-dispersion moments provide higher exploration value. Using metrics such as reward variability (Avg.~Reward Std) and intra-/inter-prompt diversity (LPIPS MPD, TCE)~\cite{kim2025test}, we compare exploration strategies centered on the Top-$K$ entropy peaks. As shown in Table~\ref{tab:reward_std_comparison}, peak-based exploration consistently achieves higher reward variance and greater generative diversity, outperforming fixed schedules. These results indicate that the $Entropy(t)$ peaks mark timesteps where the denoising trajectory is most sensitive to perturbations and thus most amenable to meaningful exploration.

Overall, these findings demonstrate that $Entropy(t)$ peaks serve as reliable indicators of critical timesteps where exploration is most beneficial.

\section{AEGPO}

Building upon the insights from Attention Entropy, we present \textbf{AEGPO}, a dual-level adaptive framework designed to improve rollout efficiency in GRPO training. AEGPO operates along two complementary dimensions:

\begin{itemize}[leftmargin=1.5em]
\item \textbf{Global Adaptive Allocation} adjusts \emph{which samples} receive more rollouts, guided by their entropy-based learning value.
\item \textbf{Local Adaptive Exploration} determines \emph{where} exploration should occur within each trajectory by identifying the timesteps with maximal attention dispersion.
\end{itemize}

Together, these components reallocate computation both across samples and across timesteps, while keeping the overall rollout budget fixed.

\begin{table*}[ht!]
\centering
\caption{Main results on \textbf{FLUX.1-dev}, \textbf{SD3.5-M}. AEGPO consistently improves performance across diverse models and frameworks (e.g., GRPO variants and DiffusionNFT). For each metric within a model block, the \textbf{highest} score is bolded, and the \underline{second-highest} score is underlined.}
\label{tab:main_results_all_models}
\begin{tabular}{@{}llcccc@{}} 
\toprule
\textbf{Base Model} & \textbf{Method} & \textbf{HPS-v2.1}$\uparrow$ & \textbf{Pick Score}$\uparrow$ & \textbf{Image Reward}$\uparrow$ & \textbf{GenEval}$\uparrow$ \\
\midrule
\multirow{5}{*}{\textbf{FLUX.1-dev}} & FLUX (Base) & 0.313 & 0.227 & 1.112 & --- \\
& DanceGRPO & 0.360 & 0.228 & 1.517 & --- \\
& \textbf{DanceGRPO + AEGPO} & \underline{0.367} & \underline{0.230} & 1.532 & --- \\
& BranchGRPO & 0.363 & 0.229 & \underline{1.603} & --- \\
& \textbf{BranchGRPO + AEGPO} & \textbf{0.374} & \textbf{0.232} & \textbf{1.624} & --- \\
\midrule
\multirow{5}{*}{\textbf{SD3.5-M}} & SD3.5-M (Base) & 0.204 & 20.51 & 0.85 & 0.63 \\
& Flow-GRPO & 0.316 & 23.50 & 1.29 & 0.88 \\
& \textbf{Flow-GRPO + AEGPO} & 0.325 & 23.59 & 1.34 & 0.90 \\
& DiffusionNFT & \underline{0.331} & \underline{23.80} & \underline{1.49} & \underline{0.94} \\
& \textbf{DiffusionNFT + AEGPO} & \textbf{0.343} & \textbf{23.88} & \textbf{1.57} & \textbf{0.95} \\
\bottomrule
\end{tabular}
\end{table*}

\subsection{Global Adaptive Allocation}
\label{sec:global_allocation}

Standard GRPO assigns a uniform number of rollouts to every prompt, ignoring the substantial variation in their learning contribution. Motivated by the sample-dependent influence observed in Sec.~\ref{sec:global_insight}, we replace static allocation with an entropy-guided, value-based mechanism.

For each sample $i$ in a batch, we compute its learning value using the sample-level entropy shift:
\[
v_i = \Delta\mathrm{Entropy}_i.
\]
A high $v_i$ indicates that the sample induces a large deviation in attention behavior and is therefore more informative, while a low $v_i$ implies diminishing returns.

AEGPO divides the batch into two tiers according to the median of $\{v_i\}$: a high-value tier and a low-value tier. Each tier receives a distinct rollout budget:
\[
r_{\text{high}} > r_{\text{low}},\qquad 
r_{\text{high}} + r_{\text{low}} = 2 r_{\text{avg}},
\]
keeping the total computation unchanged. In practice, for an average budget of 12 rollouts, we allocate $(r_{\text{low}}, r_{\text{high}}) = (8, 16)$.

\textbf{Algorithmically}, Global Adaptive Allocation can be summarized as:
\begin{enumerate}[leftmargin=1.5em]
\item Compute $v_i = \Delta\mathrm{Entropy}_i$ for each sample.
\item Split samples by the batch median of $v_i$.
\item Assign rollout budgets $\{r_{\text{low}}, r_{\text{high}}\}$ accordingly.
\end{enumerate}

This adaptive redistribution focuses computation on the samples that drive the largest policy updates, improving sample efficiency without increasing training cost.

\subsection{Local Adaptive Exploration}
\label{sec:local_exploration}

Beyond sample selection, tree-structured rollout methods require deciding \emph{when} exploration should occur within each trajectory. Existing approaches rely on fixed branching schedules, implicitly assuming that informative timesteps are shared across prompts. However, Sec.~\ref{sec:local_insight} shows that the most uncertain timesteps correspond to peaks of $Entropy(t)$, and their locations vary significantly across samples and stages, often shifting between early structural steps and late refinement phases.

To address this mismatch, AEGPO performs \textbf{entropy-guided timestep selection}. For each rollout, we compute the full entropy trajectory $\{ \mathrm{Entropy}(t) \}$ across all denoising timesteps:
\[
\mathcal{T}_{\text{peak}} = \operatorname{TopK} \big( Entropy(t) \big),
\]
where $K=4$ in all experiments. Branching is then performed exclusively at $\mathcal{T}_{\text{peak}}$.

\textbf{The procedure is:}
\begin{enumerate}[leftmargin=1.5em]
\item Compute $Entropy(t)$ for all timesteps.
\item Identify $\mathcal{T}_{\text{peak}} = \text{TopK}(Entropy(t))$.
\item Apply branching only at timesteps in $\mathcal{T}_{\text{peak}}$.
\end{enumerate}

This approach is lightweight, model-agnostic, and dynamically adapts exploration to each sample’s denoising structure. As validated in Table~\ref{tab:reward_std_comparison}, entropy-peak selection yields higher reward variance and both intra- and inter-prompt diversity compared to any fixed scheduling rule.

Overall, Global Adaptive Allocation identifies which samples deserve more computation, while Local Adaptive Exploration determines where exploration is most effective within each trajectory. Together, they form a unified, dual-level adaptive mechanism for efficient policy optimization.

\begin{figure*}[t]
    \centering
    \includegraphics[width=0.9\linewidth]{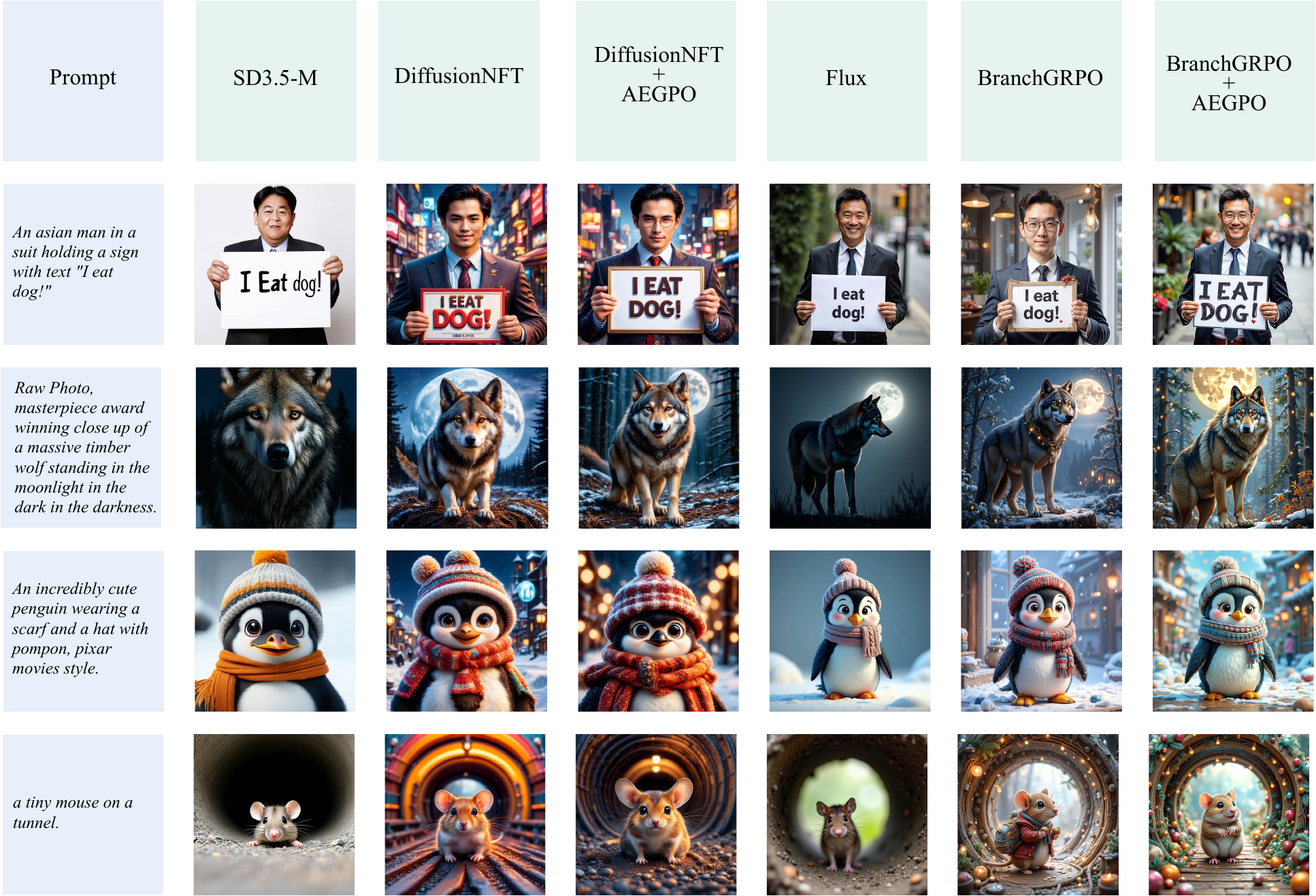}
    \caption{Qualitative comparison on challenging prompts across base models (SD3.5-M, FLUX.1-dev), baseline alignment methods (DiffusionNFT, BranchGRPO), and our AEGPO-enhanced variants. }
\label{fig:qualitative_results}
\end{figure*}

\section{Experiments}
In this section, we conduct extensive experiments to validate the effectiveness and efficiency of our proposed \textbf{AEGPO} framework. We first detail the experimental setup, including datasets, base models, and reward models (Sec.~\ref{subsec:setup}). Next, we present the main results (Sec.~\ref{subsec:main_results}), comparing AEGPO against state-of-the-art baselines across multiple models to demonstrate its superior alignment and generalizability. We follow with a detailed ablation study (Sec.~\ref{subsec:ablations}) to dissect the individual contributions of our global and local adaptive components. Finally, we provide an analysis of the computational overhead to evaluate the method's practical efficiency (Sec.~\ref{subsec:overhead}).

\subsection{Setup}
\label{subsec:setup}

\noindent \textbf{Base Models.} Our primary experiments utilize FLUX.1-Dev \cite{flux2024} and SD3.5-Medium \cite{esser2024scaling} as the main base models.

\noindent \textbf{Datasets.} We use the HPDv2.1 dataset \cite{wu2023human} and the Pick-a-Pic dataset \cite{Kirstain2023PickaPicAO} for training. For evaluation, we use the 400-prompt balanced test set from HPDv2.1 and the pickscore test set.

\noindent \textbf{Reward Models.} We employ HPSv2.1 \cite{wu2023human} as the primary reward model for training. To assess alignment across diverse human preferences, we use a comprehensive suite of reward models for evaluation, including PickScore \cite{Kirstain2023PickaPicAO}, ImageReward \cite{xu2023imagereward}, GenEval \cite{ghosh2023geneval}.

\noindent \textbf{Baselines and Implementation.} We apply AEGPO as a plug-and-play enhancement to multiple state-of-the-art frameworks, including DanceGRPO \cite{xue2025dancegrpo}, BranchGRPO \cite{li2025branchgrpo}, FlowGRPO \cite{liu2025flow}, and DiffusionNFT \cite{zheng2025diffusionnft}. In our experiments, the selected layers to calculate attention entropy for FLUX.1-dev and SD3.5-M were {5, 10, 15}. We employ a 20-step warmup period where uniform allocation is used, allowing the $\Delta\mathrm{Entropy}$ signal to become informative before activating the adaptive strategies. To ensure a fair comparison, all AEGPO-enhanced experiments are conducted under identical settings as their respective baseline papers. All GRPO-related hyperparameters are kept identical across methods.

\subsection{Main Result}
\label{subsec:main_results}
We present our main comparative results in Table \ref{tab:main_results_all_models}. The table evaluates all methods across two distinct base models (FLUX.1-dev, SD3.5-M) and multiple reward metrics. Our AEGPO framework is applied as an adaptive enhancement to DanceGRPO, BranchGRPO, and DiffusionNFT, clearly demonstrating its effectiveness and generalizability.

\noindent\textbf{Quantitative Analysis.} As shown in Table \ref{tab:main_results_all_models}, applying AEGPO consistently yields significant performance gains across all tested models and frameworks.

First, on the FLUX.1-dev model, {BranchGRPO + AEGPO} achieves state-of-the-art results, improving the HPS-v2.1 score from 0.363 to 0.374 and the Pick Score from 0.229 to 0.232 compared to the original BranchGRPO. This pattern of improvement holds for {DanceGRPO + AEGPO} as well. Second, this superiority extends beyond GRPO-style frameworks. On the SD3.5-M model, {DiffusionNFT + AEGPO} demonstrates a clear and consistent performance boost over the standard DiffusionNFT baseline. 

These results highlight a key contribution of AEGPO: its versatility. The performance improvements are consistent across both base models (FLUX.1-dev, SD3.5-M), indicating our entropy-guided approach is model-agnostic. Furthermore, its success on fundamentally different alignment frameworks—enhancing standard GRPO variants (FlowGRPO, DanceGRPO, BranchGRPO) and {DiffusionNFT} alike—strongly suggests that our adaptive strategies transfer well across different policy optimization frameworks.

\noindent\textbf{Qualitative Analysis.}
Beyond quantitative metrics, Figure \ref{fig:qualitative_results} provides a qualitative comparisons on challenging prompts across SD3.5-M, Flux..-dev, DiffusionNFT, BranchGRPO, and their AEGPO-enhanced variants. The results show that AEGPO consistently produces generations with stronger semantic alignment, improved stylistic fidelity, and more coherent visual composition. Overall, the adaptive mechanisms introduced by AEGPO lead to more reliable and prompt-faithful outputs compared to their respective baselines.

\begin{figure}
    \centering
    \includegraphics[width=0.8\linewidth]{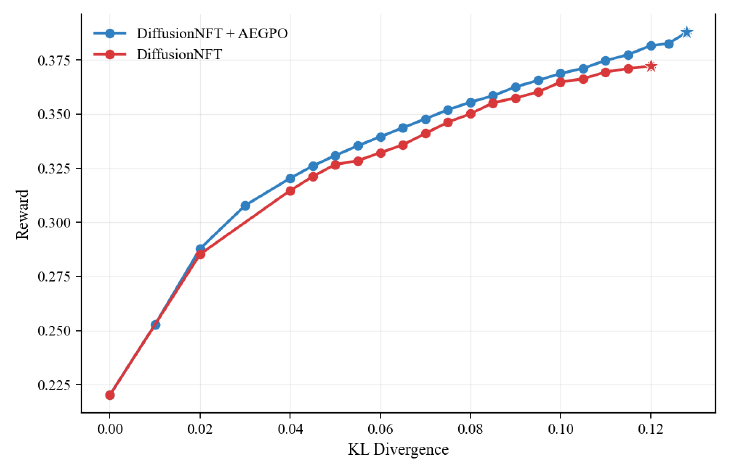}
    \caption{Reward-KL Pareto Frontier. The plot visualizes the Pareto frontier of alignment reward vs. KL divergence. The AEGPO-enhanced model (blue) consistently dominates the baseline (red), achieving a significantly higher reward for any given level of policy deviation. This indicates that AEGPO's adaptive strategy leads to a more efficient optimization path, pushing the alignment frontier to a superior ceiling.}
    \label{fig:reward_kldiv}
\end{figure}

\noindent\textbf{Reward-KL Trade-off Analysis.} 
To further understand the optimization efficiency, we analyze the reward-KL trade-off in Figure~\ref{fig:reward_kldiv}. 
The plot shows that our AEGPO-enhanced model consistently gets a better frontier the DiffusionNFT baseline, forming a superior Pareto frontier. 
This indicates that for any given level of policy deviation (KL divergence), AEGPO achieves a significantly higher reward. 
This analysis provides strong evidence that AEGPO's adaptive strategy guides the model along a fundamentally more effective and efficient optimization path.

\subsection{Ablation Study}
\label{subsec:ablations}

To disentangle the contributions of our two adaptive components, we conduct a comprehensive ablation study on the FLUX.1-dev model. We use standard BranchGRPO—featuring a fixed branching schedule and uniform sample allocation—as the baseline. We then independently activate our two modules: (1) \textit{Global Allocation} ($\mathcal{G}$), which reallocates rollouts based on $\Delta$Entropy, and (2) \textit{Local Exploration} ($\mathcal{L}$), which replaces the fixed branching schedule with our entropy-guided timestep selection.

The quantitative results are reported in Table~\ref{tab:ablation_study}. Both variants that enable only $\mathcal{G}$ or only $\mathcal{L}$ achieve clear improvements over the baseline, demonstrating that prioritizing high-value samples ($\mathcal{G}$) and adaptively selecting critical timesteps ($\mathcal{L}$) each provide independent benefits. Notably, the full AEGPO configuration ($\mathcal{G} + \mathcal{L}$) achieves the highest final reward, indicating that the two components are complementary. Jointly optimizing sample-level allocation and timestep-level exploration yields the strongest overall performance.

\begin{table}[t]
  \centering
\caption{Ablation study of AEGPO components on FLUX.1-dev, using BranchGRPO as the baseline. $\mathcal{G}$ denotes Global Adaptive Sample Allocation (Sec.~\ref{sec:global_allocation}), and $\mathcal{L}$ denotes Local Adaptive Exploration (Sec.~\ref{sec:local_exploration}), which replaces the fixed branching schedule of the baseline.}
  \label{tab:ablation_study}
  \resizebox{0.9\columnwidth}{!}{%
  \begin{tabular}{@{}lccc@{}}
    \toprule
    Method & Global ($\mathcal{G}$) & Local ($\mathcal{L}$) & HPS-v2.1 $\uparrow$ \\ 
    \midrule
    BranchGRPO (Baseline) & & & 0.363 \\
    BranchGRPO + $\mathcal{L}$ & & \checkmark & 0.369 \\
    BranchGRPO + $\mathcal{G}$ & \checkmark & & 0.371 \\
    \textbf{AEGPO (Full, $\mathcal{G} + \mathcal{L}$)} & \checkmark & \checkmark & \textbf{0.374} \\
    \bottomrule
  \end{tabular}
  } 
\end{table}

\subsection{Computational Overhead Analysis}
\label{subsec:overhead}

AEGPO introduces a modest computational overhead by extracting intermediate attention maps (from layers 5, 10, 15 ). On the FLUX.1-dev model, this adds 52 seconds to the 469 seconds baseline DanceGRPO step (an 11.1\% time increase) and raises peak VRAM from 33.5GB to 34.5GB, which is a negligible increment.

However, this per-step cost is overwhelmingly justified by the dramatic improvement in overall training efficiency. As visualized in Figure~\ref{fig:teaser}, AEGPO functions as a powerful convergence accelerator: it achieves the baseline's final reward 2$\times$ faster on DanceGRPO (in 150 steps) and 5$\times$ faster on DiffusionNFT. This demonstrates a superior time-to-performance trade-off.

\section{Discussion}
While AEGPO provides a lightweight and broadly applicable adaptive mechanism, it also has natural limitations. 
First, AEGPO introduces a small computational overhead due to extracting intermediate attention maps. 
Second, its dynamic allocation and branching strategies require a reasonably flexible and robust RL pipeline for diffusion/flow models; adopting AEGPO may therefore involve moderate engineering adjustments to existing infrastructures. 

We also examine whether other metrics could substitute attention entropy for guiding policy optimization.
Global measures such as KL divergence quantify overall model drift but provide no prompt-specific learning signal, making them unsuitable for sample-level allocation.
Stepwise proxies based on reward variance estimated via ODE/SDE solvers (e.g., TempFlow-GRPO) capture timestep sensitivity but require heavy additional forward passes.
In contrast, attention entropy remains both prompt-resolved and computationally lightweight, offering a indicator for identifying valuable samples and critical timesteps.

\section{Conclusion}

In this paper, we address inefficient static sampling in GRPO by identifying Attention Entropy as a dual-signal proxy for sample value ($\Delta$Entropy) and critical timesteps ($Entropy(t)$), proposing the adaptive AEGPO framework. Extensive experiments demonstrate AEGPO is a generalizable plug-and-play module that significantly accelerates convergence and achieves superior alignment across multiple models and policy optimization frameworks.

\clearpage

{
    \small
    \bibliographystyle{ieeenat_fullname}
    \bibliography{main}
}


\end{document}